%% file: main.tex
\documentclass{article}
\input{packages}

\def\L{{\cal L}}
\input{newcommands.tex}

\title{Binary Morphological Neural Network}
\newif\ifnames
\namestrue

\ifnames

\name{Theodore Aouad$^1$, 
Hugues Talbot$^1$
}

\address{$^1$CentraleSup\'elec, Universit\'e Paris-Saclay, Inria. Gif-sur-Yvette, France}

\else

\name{******
}

\address{******}

\fi

\begin{document}
%
\maketitle
\begin{abstract}

In the last ten years, Convolutional Neural Networks (CNNs) have formed the basis of deep-learning architectures for most computer vision tasks. However, they are not necessarily optimal. For example, mathematical morphology is known to be better suited to deal with binary images. In this work, we create a morphological neural network that handles binary inputs and outputs. We propose their construction inspired by CNNs to formulate layers adapted to such images by replacing convolutions with erosions and dilations. We give explainable theoretical results on whether or not the resulting learned networks are indeed morphological operators. We present promising experimental results designed to learn basic binary operators, and we have made our code publicly available online.

\end{abstract}
\section{Introduction}
\label{sec:intro}


Convolutional Neural Networks (CNNs) constitute the basis of most deep-learning architectures. These can learn complex task-specific processes while requiring only input-output pairs, albeit sometimes in large numbers. Since their inception in the mid-1990s, they have achieved outstanding results in computer vision and have become the go-to technology for many computer vision tasks, provided enough annotated data is available.

However, standard literature on image processing states that some tasks remain for which convolutions are not optimal. Mathematical Morphology (MM) \cite{serra1982image} is one of these. For many applications, MM operators are more suitable than convolution-based methods, particularly when dealing with binary or discrete images. However, finding the right sequence of operations and the right structuring elements can be difficult and time-consuming depending on the problem at hand~\cite{pierres_book03}. Our objective is to mimic the way CNNs are built on convolutional filters and create a morphological network that can learn a compact sequence of operators together with their optimal parameters. Morphological networks can also be used conjointly with CNNs to learn the morphological operators that would be otherwise manually engineered, as in~\cite{shit_cldice_2021}.

Learning morphological operators is not new, whereas the trend of replacing the convolution of CNNs with morphological operations is recent. Some researchers have investigated the use of the max-plus algebra \cite{mondal2020image, franchi2020deep}, for example, to perform image filtering (de-raining and de-hazing) \cite{mondal2019learning}. Others have replaced the non-differentiable max / min operators by differentiable approximations, e.g. the adaptative morphological layer \cite{shen2019deep}, the PConv layer \cite{masci2013learning}, or even the $\LMorph$ and $\SMorph$ layers \cite{kirszenberg2021going}. All these methods were studied in the context of grey-scale morphology.

In this work, we seek to learn binary morphological operators from binary image inputs. End-to-end learning of these operators could be helpful in shape analysis. We first introduce the Binary Structuring Element (BiSE) neuron, which can learn erosion and dilation together with a structuring element. The BiSE neuron is built using convolution and benefits from the highly optimized implementations of this operation. 
By stacking multiple BiSE, we build a Binary Morphological Neural Network (BiMoNN). We give theoretical explainability of the BiSE such that each learned morphological operator can be recovered. Binarizing these networks can lead to faster and cost-efficient deep networks for inference~\cite{hubara2016binarized,kim2016bitwise,simons2019review}.

Our code is publicly available online at \url{https://github.com/TheodoreAouad/Bimonn\_ICIP2022}.

\section{Method}
\label{sec:method}

\subsection{Mathematical Morphology}

Mathematical morphology \cite{serra1982image} was created to study porous materials. It is based on set theory and is well suited to studying binary images. An image of dimension $d$ ($d=2$ for 2D images or $d=3$ for 3D images) is seen as a subset of $\Z^d$: $I \subset \Z^d$. Morphological operations transform $I$ based on a small structuring element $S \subset \Z^d$. The two fundamental operations are dilation and erosion.

\begin{definition}
The \textit{dilation} ($\dil{}{}$) and \textit{erosion} ($\ero{}{}$) of an image $I \subset \Z^d$ by a structuring element $S \subset \Z^d$ are defined as:
\begin{align}
    \delta_S(I) =  \dil{I}{S} &= \bigcup_{s \in S}{(I + s)}\\
    \varepsilon_S(I) = \ero{I}{\check{S}} &= \bigcap_{s \in \check{S}}{(I + s)},
\end{align}
where $\check{S}$ is the symmetric of $S$ with respect to the origin.
\end{definition}

These operators thus defined are \emph{adjunct}, i.e. $\forall I,J,S, I\subseteq \varepsilon_S(J) \Longleftrightarrow \delta_S(I) \subseteq J$. All basic morphological operations are obtained from erosions and dilations. The opening is the application of erosion followed by its adjunct dilation, and the closing is the other way around. Here we consider any composition of dilations or erosions, extending to any sequence of openings or closings.

Erosions and dilations are similar to convolutions: the structuring elements can be interpreted as the kernel for the convolutional filter, and the sum is replaced by max or min operator. Therefore, we must establish a model that can learn the structuring element and operation type: erosion or dilation.

\subsection{Binary Structuring Element Neuron}

We now define the BiSE (Binary Structuring Element) neuron, which can learn both the operation and the structuring element. The BiSE neuron is built upon the convolution operation. First, we notice that the dilation and erosion can be exactly expressed using convolution. Note that in practice, for the erosion, we learn the symmetric $\varepsilon_{\check{S}}$ such that $\varepsilon_{\check{S}}(I) = \ero{I}{S}$.

\begin{proposition}[Morphological operators from convolution]\label{prop:conv-morp}
Let $S \subset \Z^d$ be a binary structuring element and $X \subset \Z^d$ be a binary image. 
  \begin{align}
  \dil{X}{S} &= \bigg(\conv{\indicator{X}}{\indicator{S}} \geq 1\bigg)\\   
  \ero{X}{S} &= \bigg(\conv{\indicator{X}}{\indicator{S}} \geq \card{S}\bigg)
  \end{align}
  
\end{proposition}

For the same structuring element $S$, the difference between dilation and erosion is determined by a scalar. Learning the operation is the same as learning this scalar.

To learn $S \subset \Z^d$, first we suppose that $S$ is bounded: $S \subseteq \Omega$ with $\card{\Omega} < +\infty$. Let $\Omega$ be the grid bounded by some integer $n$, $\Omega = \Z^d \cap [-n, n]^d$. Similarly to~\cite{kim2016bitwise}, we define a relaxed weight $W \subset \R^{\Omega}$, we apply a smooth increasing threshold function $\xi$ such that $\xi(W(i)) \approx 1$ if $i \in S$, else $\xi(W(i)) \approx 0$. We use the hyperbolic tangent: 

\begin{equation}
    \xi(x) = \frac{1}{2} \tanh(x) + \frac{1}{2}
\end{equation}

We introduce the softplus function $f^+: x \in \R \mapsto \ln(1 + \exp (x)) + 0.5$ to ensure that $f^+(x) > 0.5$.

 

\begin{definition}[BiSE neuron]
Let $W \in \R^{\Omega}$ be a weight matrix, $b \in \R$ a bias and $p \in \R$ a scaling number. We define a \textbf{BiSE (Binary Structuring Element) neuron} as follow:
\begin{equation}
\bise_{W, b, p}: x \in [0, 1]^{\Z^d} \mapsto \xi(p (\conv{x}{\xi(W)} - f^+(b))) \in [0, 1]^{\Z^d}
\end{equation}        
\end{definition}

First, the weights are thresholded. Then we apply the morphological operation: we perform the convolution and subtract a bias. The bias is forbidden to become negative (in practice, $f^+(b) > 0.5$): we avoid the bias at $0$, which leads to constant output and zero-grad zones. Finally, we have to threshold this result. Before thresholding, the result is multiplied by a scaling factor $p$.

The BiSE neuron can learn erosion, dilation, and the associated structuring element. The weights $W$ learn the structuring element, and the bias $b$ determines the operation. The scaling number $p$ has two purposes. It determines how close to binary the output is, and if $p < 0$, the output is inverted, so in theory, we could learn the complementation as well.

Our thresholding function, the hyperbolic tangent, is smooth to allow back-propagation; we do not deal with binary images. We thus introduce the concept of almost binary images, which are more flexible and are easier to handle than binary images.

\begin{definition}[Almost Binary Image]
We say an image $I \in [0, 1]^{\Z^d}$ is \textbf{almost binary} if there exists $u < v \in [0, 1]$ such that $I(\Z^d) \cap ]u, v[ = \emptyset$. We denote this set $\almostbinary{u}{v}$.
\end{definition}

A pixel value of an almost binary image is either close to 0 or close to 1.

When is a BiSE neuron equivalent to dilation or erosion by a structuring element $S$? The following propositions allow us to perform a verification for a given structuring element. Given the weights $W$ and the bias $b$, we can check if, in its current state, the BiSE neuron is a dilation or erosion by this structuring element.

\begin{proposition}[Dilation / Erosion Equivalence] \label{prop:check-dila}
    We assume the weights are thresholded: $W \in [0, 1]^{\Omega}$. Given an almost binary input in $\almostbinary{u}{v}$\\
    
\begin{itemize}
    \item $\epsilon_{W, b, +\infty}$ is a dilation by $S$ if and only if 
    \begin{equation}
        \sum_{i \in \Omega \backslash S}{w_i} + u \sum_{i \in S}{w_i} \leq b < v \min_{i \in S}{w_i}
    \end{equation}
    \item$\epsilon_{W, b, +\infty}$ is an erosion by $S$ if and only if 
    \begin{equation}
        \max_{j \in S}\bigg( \sum_{i \in  S \backslash j}{(w_i)}  + u \cdot w_j\bigg)  \leq b < v\sum_{i \in S}{W_i}
    \end{equation}
\end{itemize}

If either of these expressions is fulfilled, we say that the BiSE neuron is \textbf{activated}.

\end{proposition}


If the weights and bias are correctly learned, the structuring element can be recovered by thresholding the weights for some value. The suitable threshold is given in proposition \ref{prop:linear}.

\begin{proposition}[Linear Check]\label{prop:linear}

Let us assume the BiSE is activated for almost binary images $\almostbinary{u}{v}$.  Let $b$ be the BiSE bias, let $W$ be the normalized weights ($W \in ]0, 1[^{\Omega}$). 
Then there exists $\tau \in \R$ such that $S = \{i \in \Omega ~|~  W(i) \geq \tau\}$.

\begin{itemize}
    \item If the BiSE is a dilation 
\begin{equation}
\tau = \frac{b}{v}
\end{equation}
    \item If the BiSE is an erosion
\begin{equation}
\tau = \frac{\sum_{k \in \Omega}{W_k} - b}{1 - u}    
\end{equation}
\end{itemize}

\end{proposition}

To find out the learned structuring element and operation, we only need to check both thresholds. This can be done in $\mathcal{O}(\card{\Omega})$ operations.

Erosion and dilation are dual operations: applying a dilation is the same as applying an erosion to the background. This property allows us to recover the inequalities of one operation to get its dual. Finally, if the BiSE neuron is activated, then its output is almost binary. The notion of almost binary images is justified: we now deal with almost binary images instead of binary images.

\subsection{BiMoNN}

We can now define the Binary Morphological Neural Network (BiMoNN) as a composition of multiple BiSEs:
\begin{equation}
    \bimonn = \bise_L \circ ... \circ \bise_1
\end{equation}

If each BiSE is activated, the BiMoNN's inputs and outputs are almost binary. In theory, this framework can learn any sequence of dilations or erosions with any structuring element, including opening and closing.


We learn the BiMoNN using the classical deep learning framework. Given a loss $\L$, we use a masked version of $\L$. Borders act depending on our interpretation of the value out-of-bounds pixels. To avoid this problem, we mask the borders of size kernel shape divided by two.
Parameters are updated with the Adam~\cite{kingma2017adam} optimizer. We initialize the biases at $f^+(2) = 0.63$. The weights $W$ of the convolutions have kaiming uniform initialization~\cite{glorot2010understanding}.

\section{Experiments}
\label{sec:experiments}

\subsection{Datasets}


We create a dataset of generated images that we call \diskorect{}. Each image is made of random-shaped and random-oriented rectangles and disks. Then, we add some random Bernoulli noise. Finally, complementation is applied half the time. The images dimensions are $50\times 50$.



\mnist{}~\cite{deng2012mnist} is a dataset of 70000 handwritten digits of size $28\times 28$. The images are grey-level. To be able to work with structuring elements of size $(7,7)$, we reshape them to $(50,50)$, with a cubic interpolation \cite{keys1981cubic}. Then we threshold the image to recover binary images.

In order to analyze the duality of operators, we also test our network on the complementation of \mnist{}, which we call \text{inverted \mnist{}}.

\begin{figure}[h]
    \centering
    \begin{subfigure}{.3\linewidth}
        \includegraphics[width=.48\textwidth]{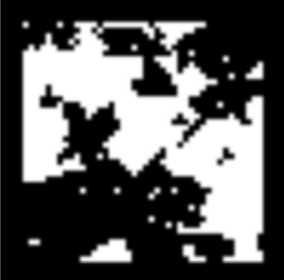} \includegraphics[width=.48\textwidth]{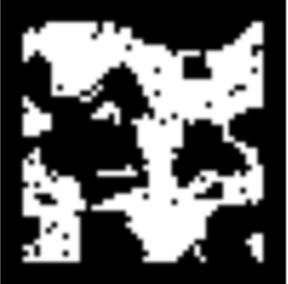}
        \caption{Diskorect}
        \label{subfig:diskorect}
    \end{subfigure}
    \begin{subfigure}{.3\linewidth}
        \includegraphics[width=.48\textwidth]{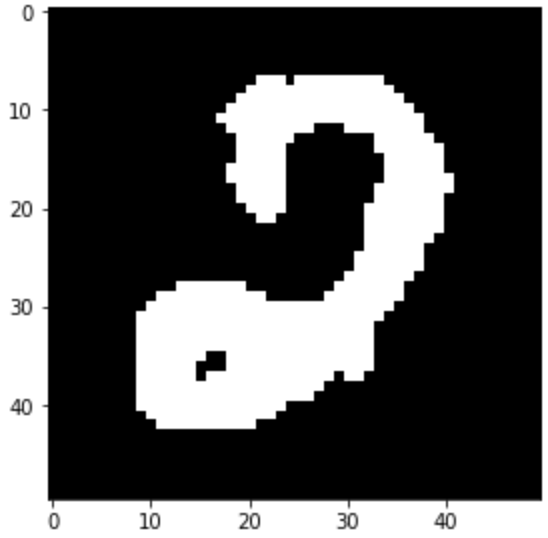} \includegraphics[width=.48\textwidth]{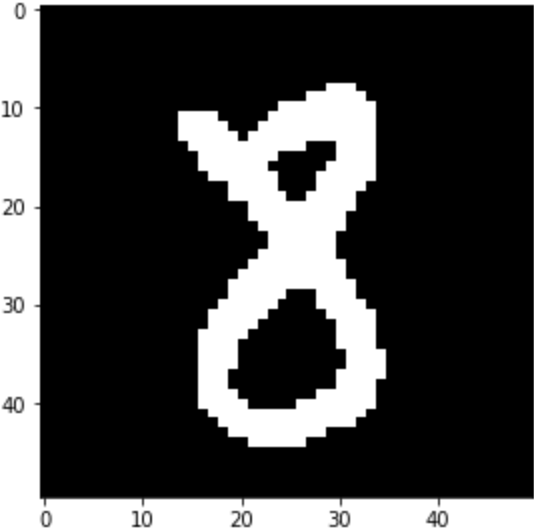}
        \caption{MNIST}
        \label{subfig:mnist}
    \end{subfigure}
    \begin{subfigure}{.3\linewidth}
        \includegraphics[width=.48\textwidth]{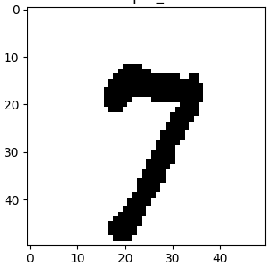} \includegraphics[width=.48\textwidth]{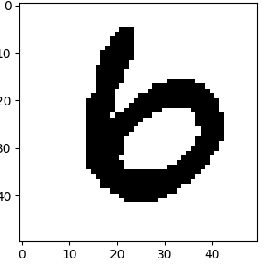}
        \caption{Inverted MNIST}
        \label{subfig:inverted_mnist}
    \end{subfigure}
    \caption{Datasets example.}
    \label{fig:dataset_example}
\end{figure}

\subsection{Experiment description}

We check the ability of a BiMoNN to learn basic morphological operators. We attempt to learn the erosion and dilation (table \ref{tab:erodila}), as well as the opening (erosion then dilation) and closing (dilation then erosion) (table \ref{tab:opeclos}). The protocol is the same for each operation: the morphological operator is applied to the input image to create the target $y_i$. To learn the erosion and dilation, we train a single BiSE neuron. To learn the opening and closing, we stack two BiSE neurons. We follow the DICE \cite{dice1945measures} of the target images vs. predicted images. We also check if each BiSE neuron is activated.

For both \mnist{} and inverted \mnist{}, the structuring elements are of size $5 \times 5$ for the erosion and dilation; the transformations are too significant compared to the size of the digits otherwise.

The parameter $p$ is fixed $p = 4$. For \diskorect{}, we use the Dice loss \cite{milletari2016vnet} with learning rate $0.01$ for the erosion / dilation and $0.001$ for the opening / closing.  For \mnist{}, we use the MSE loss with learning rate $0.1$ for the erosion / dilation and $0.01$ for the opening / closing.



\begin{table}[h]
    \centering
    \caption{Results on erosion and dilation. DICE error ($1 - \text{DICE}$) is presented for each case. \textcolor{blue}{\checkmark} indicates if the neuron is activated.}
    \resizebox{\linewidth}{!}{\input{results_table_dilero3}}    
    \label{tab:erodila}
\end{table}

\begin{table}[h]
    \centering
    \caption{Results on opening and closing. DICE error ($1 - \text{DICE}$) is presented for each case. \textcolor{blue}{\checkmark} indicates if the neuron is activated.}
    \resizebox{\linewidth}{!}{\input{results_table_opeclos3}}
    \label{tab:opeclos}
\end{table}

\section{Discussion}
\label{sec:discussion}

\subsection{Erosion and Dilation}

For any optimizer or any initialization, on both \mnist{} and \diskorect{}, the dilation is learned perfectly for all structuring elements: after a few hundred iterations, the DICE is 1 for the validation set. The BiSE are all activated except for the cross. On the \diskorect{} dataset, the BiSE neurons are activated after a few thousand iterations for the disk and stick. 

For the erosion, on \diskorect{}, we obtain similar results but with slower convergence. Sometimes the weights are darker than in the dilation case. This is not a problem: if the BiSE is activated, the weights do not need to be as high as 1. On \mnist{}, the learned disk is limited to its border, which is not surprising: given $S$ and its border $\partial S$, the difference between $\ero{I}{S}$ and $\ero{I}{\partial S}$ is small and only visible on a dataset with small holes. While the DICE is not 1, the error is small at $0.003$.

The perfect metric is reached before the BiSE neurons are activated. Indeed, it is possible for the inequality not to be fulfilled, while the BiSE is a dilation only for a specific dataset $D \subsetneq \Z^d$. Could a necessary and sufficient condition be established for a specific dataset to know if the BiSE is a dilation on this dataset? This question is left for future work.

\subsection{Opening and Closing} \label{subsec:opeclos}


On \diskorect{}, for both opening and closing, we learn perfectly the stick and cross, and almost all the BiSE are activated. However, the disk is more challenging to learn for opening (resp. closing): the DICE is still high at $0.93$ (resp. $0.96$). 

On \mnist{}, the opening is learned well for all structuring elements. However, the training on closing returns chaotic weights for the disk and cross. To explain this, we notice that closing with $S$ has little effect on the image for this dataset. Therefore many weight combinations yield a similar transformation. See the disk closing: even with totally different structuring elements, the error is small at $0.002$. For the inverted \mnist{}, the situation is almost the same, but for the dual operators: we will dive into that in the next paragraph.

We compare to $\SMorph$ and $\LMorph$~\cite{kirszenberg2021going}. On both datasets, they achieve good results on erosion and dilation, with perfect metric and good structuring elements. However, they fail to converge on the opening and closing and present stability issues (because of an exponential in their formulation). This is not surprising as they are both built on a differentiable softmax function that is not suitable to deal with binary elements.


\subsection{A few words on duality}

In mathematical morphology, two operators $\delta$ and $\varepsilon$ are dual if $\forall X ~,~ \overline{\delta(X)} = \varepsilon(\overline{X})$. This is the case for the couples (dilation, erosion) and (opening, closing). Let $D$ be a dataset and $\overline{D}$ the set of its complementations. We formulate the hypothesis $\mathcal{H}$: are the training of $\delta$ on $D$ and the training of $\varepsilon$ on $\overline{D}$ similar?

Let $D$ be the \diskorect{} dataset. By construction, $\overline{D} = D$ and the training of dual operators should be the same. For the erosion and dilation, the results are not exactly the same. For the erosion, the structuring elements are darker, and the model takes longer to converge: from 10 times slower on \diskorect{} to a hundred times slower on \mnist{}. We still manage to learn both operations. On the other hand, the opening and closing do not behave the same on the closing, with dissimilar weights. These results contradict $\mathcal{H}$.

Now, let $D$ be the \mnist{} dataset. The dilation behaves identically for both $D$ and $\overline{D}$, and the same can be said for the erosion, which goes against $\mathcal{H}$. However, even though the learned weights seemed chaotic, the closing on $D$ works similarly to the opening on $\overline{D}$, which corroborates $\mathcal{H}$. This similarity also happens for the opening on $D$ and closing on $\overline{D}$, being only different for the disk.

The case of \mnist{} and its invert suggests that there is a strong link between the learning of dual operators. However, they do not learn exactly the same. We leave the study of this link to future work.

\section{Conclusion}
\label{sec:print}

We created a neural network built to operate on binary images. We successfully learn some morphological operations: results on dilations and erosions are perfect. Overall, we achieved good results on the opening and closing, only failing for the disk and more complicated operations. We also provide explainability results to understand the operation of each neuron.

We sometimes reach a perfect metric without the neurons being theoretically morphological operators. This raises the need for more relaxed inequalities depending on the dataset properties. Enforcing that each BiSE is activated (e.g., using a regularization loss) would allow us to binarize the entire network.

To complete our network, we should also learn the complementation and multiple filters (e.g., to learn the top-hat operators). The complementation can be done by learning the parameter $P$ and allowing it to be negative. While the BiSE could theoretically complement its results, it was not yet demonstrated in practice:  we leave it to future work. To fully simulate the CNNs' multiple-filter structure, we will incorporate intersection and union of filters. 

These results demonstrate that we can learn simple morphological operators. In the near future, we will investigate its integration in more complex pipelines to leverage useful morphological information, with a view to construct high-performing binary deep networks.

\bibliographystyle{IEEEbib}
\typeout{}
\bibliography{refs}

\pagebreak

\begin{appendices}
\input{annexe}
\input{appendix}

\end{appendices}

\end{document}

%% file: packages.tex
\usepackage[toc,page]{appendix}

\usepackage{caption}
\usepackage{subcaption}
\usepackage[preprint]{spconf}
\usepackage{amsmath,graphicx,amssymb,mathabx}
\usepackage{hyperref}
\usepackage{enumitem}
\setlist{nosep, leftmargin=14pt}

\usepackage{mwe} 
\usepackage[ruled,vlined]{algorithm2e}
\usepackage[utf8]{inputenc}
\usepackage[most]{tcolorbox}
\usepackage{bbm}
\usepackage{theorem}
\usepackage[export]{adjustbox}
\usepackage{xcolor}
\usepackage{multicol}

%% file: newcommands.tex
\newcommand{\discint}[2]{[\![ #1, #2 ]\!]}

\newcommand{\card}[1]{|#1|}

\newcommand{\R}{\mathbb{R}}

\newcommand{\cC}{\mathcal{C}}
\newcommand{\N}{\mathbb{N}}
\newcommand{\Z}{\mathbb{Z}}

\newcommand{\I}{\mathcal{I}}

\newcommand{\dil}[2]{#1 \oplus #2}
\newcommand{\ero}[2]{#1 \ominus #2}

\newcommand{\indicator}[1]{\mathbbm{1}_{#1}}
\newcommand{\conv}[2]{#1 \circledast #2}

\newcommand{\bise}{\epsilon}
\newcommand{\lui}{LUI}
\newcommand{\bisel}{\phi}
\newcommand{\bimonn}{\ensuremath{\text{BiMoNN}}}

\newcommand{\almostbinary}[2]{\I(#1, #2)}
\newcommand{\mnist}{MNIST}
\newcommand{\diskorect}{Diskorect}
\newcommand{\SMorph}{\ensuremath{\mathcal{S}\textit{Morph}}}
\newcommand{\LMorph}{\ensuremath{\L\textit{Morph}}}

\theoremstyle{plain}
\newtheorem{proposition}{Proposition}
\newtheorem{definition}{Definition}

%% file: results_table_dilero3.tex
     \begin{tabular}{ | c | c | c | c | c | c | c | c | c | c |}
        \hline
        Dataset & \multicolumn{3}{c|}{Diskorect}  & \multicolumn{3}{c|}{MNIST} & \multicolumn{3}{c|}{Inverted MNIST} \\
        \hline
        Operation & Disk & Stick & Cross & Disk & Stick & Cross  & Disk & Stick & Cross \\
        \hline
        Target
        & \begin{minipage}{.12\linewidth}
          \includegraphics[width=\textwidth,left]{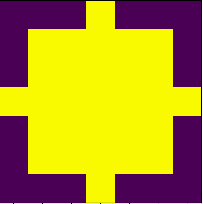}
        \end{minipage}
        & \begin{minipage}{.12\linewidth}
          \includegraphics[width=\textwidth]{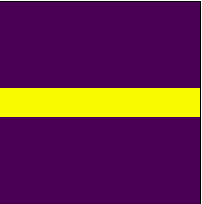}
        \end{minipage}
        & \begin{minipage}{.12\linewidth}
          \includegraphics[width=\textwidth]{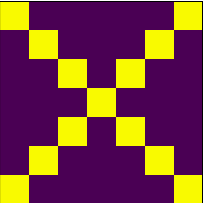}
        \end{minipage}
        & \begin{minipage}{.12\linewidth}
          \includegraphics[width=\textwidth]{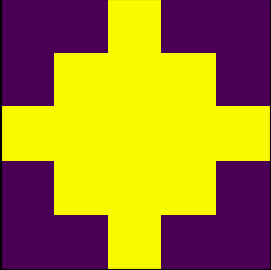}
        \end{minipage}
        & \begin{minipage}{.12\linewidth}
          \includegraphics[width=\textwidth]{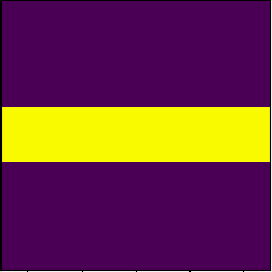}
        \end{minipage}
        & \begin{minipage}{.12\linewidth}
          \includegraphics[width=\textwidth]{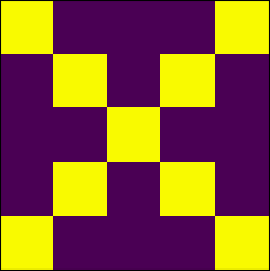}
        \end{minipage}
        & \begin{minipage}{.12\linewidth}
          \includegraphics[width=\textwidth]{selem_results2/true_disk5.png}
        \end{minipage}
        & \begin{minipage}{.12\linewidth}
          \includegraphics[width=\textwidth]{selem_results2/true_hstick5.png}
        \end{minipage}
        & \begin{minipage}{.12\linewidth}
          \includegraphics[width=\textwidth]{selem_results2/true_dcross5.png}
        \end{minipage}
        \\        
        \hline
        Dilation $\dil{}{}$
        & \hspace{-0.4cm}
        \begin{subfigure}{.12\linewidth}
          \includegraphics[width=\textwidth]{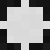}\textcolor{blue}{\checkmark}
          
          \vspace{-0.6cm}
          \center{{\tiny ${0.000}$}}
        \end{subfigure}
        & \hspace{-0.4cm}
        \begin{subfigure}{.12\linewidth}
          \includegraphics[width=\textwidth]{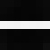}\textcolor{blue}{\checkmark}
          
          \vspace{-0.6cm}
          \center{{\tiny ${0.000}$}}
        \end{subfigure}
        & \hspace{-0.4cm}
        \begin{subfigure}{.12\linewidth}
          \includegraphics[width=\textwidth]{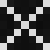}\textcolor{red}{$\times$}
          
          \vspace{-0.6cm}
          \center{{\tiny ${0.000}$}}
        \end{subfigure}
        & \hspace{-0.4cm}
        \begin{subfigure}{.12\linewidth}
          \includegraphics[width=\textwidth]{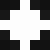}\textcolor{red}{$\times$}
          
          \vspace{-0.6cm}
          \center{{\tiny ${0.000}$}}
        \end{subfigure}
        & \hspace{-0.4cm}
        \begin{subfigure}{.12\linewidth}
          \includegraphics[width=\textwidth]{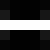}\textcolor{blue}{\checkmark}
          
          \vspace{-0.6cm}
          \center{{\tiny ${0.000}$}}
        \end{subfigure}
        & \hspace{-0.4cm}
        \begin{subfigure}{.12\linewidth}
          \includegraphics[width=\textwidth]{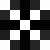}\textcolor{red}{$\times$}
          
          \vspace{-0.6cm}
          \center{{\tiny ${0.000}$}}
        \end{subfigure}
                & \hspace{-0.4cm}
                \begin{subfigure}{.12\linewidth}
          \includegraphics[width=\textwidth]{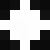}\textcolor{red}{$\times$}
          
          \vspace{-0.6cm}
          \center{{\tiny ${0.000}$}}
        \end{subfigure}
        & \hspace{-0.4cm}
        \begin{subfigure}{.12\linewidth}
          \includegraphics[width=\textwidth]{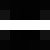}\textcolor{blue}{\checkmark}
          
          \vspace{-0.6cm}
          \center{{\tiny ${0.000}$}}
        \end{subfigure}
        & \hspace{-0.4cm}
        \begin{subfigure}{.12\linewidth}
          \includegraphics[width=\textwidth]{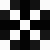}\textcolor{red}{$\times$}
          
          \vspace{-0.6cm}
          \center{{\tiny ${0.000}$}}
        \end{subfigure}
        
        \\       
        \hline
        Erosion $\ero{}{}$
        & \hspace{-0.4cm}
        \begin{subfigure}{.12\linewidth}
          \includegraphics[width=\textwidth]{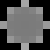}\textcolor{blue}{\checkmark}
          
          \vspace{-0.6cm}
          \center{{\tiny ${0.000}$}}
        \end{subfigure}
        & \hspace{-0.4cm}
        \begin{subfigure}{.12\linewidth}
          \includegraphics[width=\textwidth]{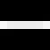}\textcolor{blue}{\checkmark}
          
          \vspace{-0.6cm}
          \center{{\tiny ${0.000}$}}
        \end{subfigure}
        & \hspace{-0.4cm}
        \begin{subfigure}{.12\linewidth}
          \includegraphics[width=\textwidth]{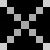}\textcolor{blue}{\checkmark}
          
          \vspace{-0.6cm}
          \center{{\tiny ${0.000}$}}
        \end{subfigure}
        & \hspace{-0.4cm}
        \begin{subfigure}{.12\linewidth}
          \includegraphics[width=\textwidth]{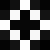}\textcolor{red}{$\times$}
          
          \vspace{-0.6cm}
          \center{{\tiny ${0.000}$}}
        \end{subfigure}
        & \hspace{-0.4cm}
        \begin{subfigure}{.12\linewidth}
          \includegraphics[width=\textwidth]{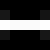}\textcolor{blue}{\checkmark}
          
          \vspace{-0.6cm}
          \center{{\tiny ${0.000}$}}
        \end{subfigure}
        & \hspace{-0.4cm}
        \begin{subfigure}{.12\linewidth}
          \includegraphics[width=\textwidth]{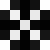}\textcolor{red}{$\times$}
          
          \vspace{-0.6cm}
          \center{{\tiny ${0.000}$}}
        \end{subfigure}
        & \hspace{-0.4cm}
        \begin{subfigure}{.12\linewidth}
          \includegraphics[width=\textwidth]{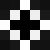}\textcolor{red}{$\times$}
          
          \vspace{-0.6cm}
          \center{{\tiny ${0.000}$}}
        \end{subfigure}
        & \hspace{-0.4cm}
        \begin{subfigure}{.12\linewidth}
          \includegraphics[width=\textwidth]{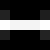}\textcolor{blue}{\checkmark}
          
          \vspace{-0.6cm}
          \center{{\tiny ${0.000}$}}
        \end{subfigure}
        & \hspace{-0.4cm}
        \begin{subfigure}{.12\linewidth}
          \includegraphics[width=\textwidth]{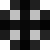}\textcolor{red}{$\times$}
          
          \vspace{-0.6cm}
          \center{{\tiny ${0.000}$}}
        \end{subfigure}
        \\
        \hline
      \end{tabular}

%% file: results_table_opeclos3.tex
     \begin{tabular}{ | c | c | c | c | c | c | c | c | c | c |}
        \hline
        Dataset & \multicolumn{3}{c|}{Diskorect}  & \multicolumn{3}{c|}{MNIST} & \multicolumn{3}{c|}{Inverted MNIST} \\
        \hline
        Operation & Disk & Stick & Cross & Disk & Stick & Cross & Disk & Stick & Cross \\
        \hline
        Target
        & \begin{minipage}{.12\linewidth}
          \includegraphics[width=\textwidth]{selem_results/true_disk7.png}
        \end{minipage}
        & \begin{minipage}{.12\linewidth}
          \includegraphics[width=\textwidth]{selem_results/true_hstick7.png}
        \end{minipage}
        & \begin{minipage}{.12\linewidth}
          \includegraphics[width=\textwidth]{selem_results/true_dcross7.png}
        \end{minipage}
        & \begin{minipage}{.12\linewidth}
          \includegraphics[width=\textwidth]{selem_results/true_disk7.png}
        \end{minipage}
        & \begin{minipage}{.12\linewidth}
          \includegraphics[width=\textwidth]{selem_results/true_hstick7.png}
        \end{minipage}
        & \begin{minipage}{.12\linewidth}
          \includegraphics[width=\textwidth]{selem_results/true_dcross7.png}
        \end{minipage}
        & \begin{minipage}{.12\linewidth}
          \includegraphics[width=\textwidth]{selem_results/true_disk7.png}
        \end{minipage}
        & \begin{minipage}{.12\linewidth}
          \includegraphics[width=\textwidth]{selem_results/true_hstick7.png}
        \end{minipage}
        & \begin{minipage}{.12\linewidth}
          \includegraphics[width=\textwidth]{selem_results/true_dcross7.png}
        \end{minipage}
        \\
        \hline
        Opening $\circ$
        & \hspace{-0.4cm}
        \begin{subfigure}{.12\linewidth}
          \includegraphics[width=\textwidth]{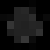}\textcolor{red}{$\times$}
          \includegraphics[width=\textwidth]{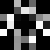}\textcolor{red}{$\times$}
          \vspace{-1cm}\center{{\tiny ${0.072}$}}
        \end{subfigure}
        & \hspace{-0.4cm}
        \begin{subfigure}{.12\linewidth}
          \includegraphics[width=\textwidth]{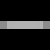}\textcolor{blue}{\checkmark}
          \includegraphics[width=\textwidth]{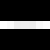}\textcolor{blue}{\checkmark}
          \vspace{-1cm}\center{{\tiny ${0.000}$}}
        \end{subfigure}
        & \hspace{-0.4cm}
        \begin{subfigure}{.12\linewidth}
          \includegraphics[width=\textwidth]{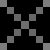}\textcolor{blue}{\checkmark}
          \includegraphics[width=\textwidth]{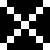}\textcolor{blue}{\checkmark}
          \vspace{-1cm}\center{{\tiny ${0.002}$}}
        \end{subfigure}
        & \hspace{-0.4cm}
        \begin{subfigure}{.12\linewidth}
          \includegraphics[width=\textwidth]{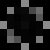}\textcolor{red}{$\times$}
          \includegraphics[width=\textwidth]{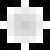}\textcolor{blue}{\checkmark}
          \vspace{-1cm}\center{{\tiny ${0.008}$}}
        \end{subfigure}
        & \hspace{-0.4cm}
        \begin{subfigure}{.12\linewidth}
          \includegraphics[width=\textwidth]{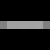}\textcolor{blue}{\checkmark}
          \includegraphics[width=\textwidth]{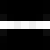}\textcolor{blue}{\checkmark}
          \vspace{-1cm}\center{{\tiny ${0.000}$}}
        \end{subfigure}
        & \hspace{-0.4cm}
        \begin{subfigure}{.12\linewidth}
          \includegraphics[width=\textwidth]{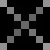}\textcolor{red}{$\times$}
          \includegraphics[width=\textwidth]{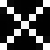}\textcolor{blue}{\checkmark}
          \vspace{-1cm}\center{{\tiny ${0.001}$}}
        \end{subfigure}
        & \hspace{-0.4cm}
        \begin{subfigure}{.12\linewidth}
          \includegraphics[width=\textwidth]{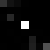}\textcolor{red}{$\times$}
          \includegraphics[width=\textwidth]{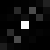}\textcolor{red}{$\times$}
          \vspace{-1cm}\center{{\tiny ${0.006}$}}
        \end{subfigure}
        & \hspace{-0.4cm}
        \begin{subfigure}{.12\linewidth}
          \includegraphics[width=\textwidth]{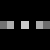}\textcolor{blue}{\checkmark}
          \includegraphics[width=\textwidth]{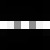}\textcolor{red}{$\times$}
          \vspace{-1cm}\center{{\tiny ${0.001}$}}
        \end{subfigure}
        & \hspace{-0.4cm}
        \begin{subfigure}{.12\linewidth}
          \includegraphics[width=\textwidth]{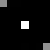}\textcolor{red}{$\times$}
          \includegraphics[width=\textwidth]{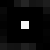}\textcolor{red}{$\times$}
          \vspace{-1cm}\center{{\tiny ${0.012}$}}
        \end{subfigure}
        \\
        \hline
        Closing $\bullet$
        & \hspace{-0.4cm}
        \begin{subfigure}{.12\linewidth}
          \includegraphics[width=\textwidth]{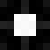}\textcolor{red}{$\times$}
          \includegraphics[width=\textwidth]{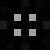}\textcolor{red}{$\times$}
          \vspace{-1cm}\center{{\tiny ${0.038}$}}
        \end{subfigure}
        & \hspace{-0.4cm}
        \begin{subfigure}{.12\linewidth}
          \includegraphics[width=\textwidth]{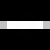}\textcolor{blue}{\checkmark}
          \includegraphics[width=\textwidth]{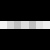}\textcolor{blue}{\checkmark}
          \vspace{-1cm}\center{{\tiny ${0.000}$}}
        \end{subfigure}
        & \hspace{-0.4cm}
        \begin{subfigure}{.12\linewidth}
          \includegraphics[width=\textwidth]{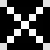}\textcolor{blue}{\checkmark}
          \includegraphics[width=\textwidth]{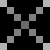}\textcolor{red}{$\times$}
          \vspace{-1cm}\center{{\tiny ${0.000}$}}
        \end{subfigure}
        & \hspace{-0.4cm}
        \begin{subfigure}{.12\linewidth}
          \includegraphics[width=\textwidth]{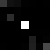}\textcolor{red}{$\times$}
          \includegraphics[width=\textwidth]{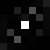}\textcolor{red}{$\times$}
          \vspace{-1cm}\center{{\tiny ${0.009}$}}
        \end{subfigure}
        & \hspace{-0.4cm}
        \begin{subfigure}{.12\linewidth}
          \includegraphics[width=\textwidth]{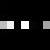}\textcolor{blue}{\checkmark}
          \includegraphics[width=\textwidth]{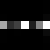}\textcolor{red}{$\times$}
          \vspace{-1cm}\center{{\tiny ${0.001}$}}
        \end{subfigure}
        & \hspace{-0.4cm}
        \begin{subfigure}{.12\linewidth}
          \includegraphics[width=\textwidth]{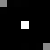}\textcolor{red}{$\times$}
          \includegraphics[width=\textwidth]{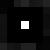}\textcolor{red}{$\times$}
          \vspace{-1cm}\center{{\tiny ${0.020}$}}
        \end{subfigure}
        & \hspace{-0.4cm}
        \begin{subfigure}{.12\linewidth}
          \includegraphics[width=\textwidth]{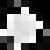}\textcolor{red}{$\times$}
          \includegraphics[width=\textwidth]{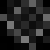}\textcolor{red}{$\times$}
          \vspace{-1cm}\center{{\tiny ${0.009}$}}
        \end{subfigure}
        & \hspace{-0.4cm}
        \begin{subfigure}{.12\linewidth}
          \includegraphics[width=\textwidth]{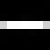}\textcolor{blue}{\checkmark}
          \includegraphics[width=\textwidth]{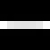}\textcolor{blue}{\checkmark}
          \vspace{-1cm}\center{{\tiny ${0.000}$}}
        \end{subfigure}
        & \hspace{-0.4cm}
        \begin{subfigure}{.12\linewidth}
          \includegraphics[width=\textwidth]{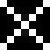}\textcolor{blue}{\checkmark}
          \includegraphics[width=\textwidth]{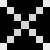}\textcolor{blue}{\checkmark}
          \vspace{-1cm}\center{{\tiny ${0.000}$}}
        \end{subfigure}
        \\
        \hline
      \end{tabular}

%% file: annexe.tex
\section{Usage of duality for erosion and dilation equivalence} \label{annex:inequality-duality}

Erosion and dilation are dual operations, meaning $\forall X, S \in \{0, 1\}^{\Omega}, \dil{X}{S} = \overline{\ero{\bar{X}}{S}}$: applying a dilation is the same as applying an erosion to the background. This property allows us to recover the inequalities of the dilation using those of the erosions, and the other around. Let us show this in the binary case.

Let us consider two BiSE layers $\bise^1_{W, b_1}$ and $\bise^2_{W, b_2}$ sharing the same weights, but the first one being a dilation and the second one being an erosion. Let us denote $u_e, v_e$ the bounds for the erosion and $u_d, v_d$ the bounds for the dilation. We have:

\begin{align}
    u_d = \sup_{X \in \{0, 1\}^{\Omega}, i \in \Omega}\{\conv{\indicator{X}}{W}(i) ~|~ i \notin \dil{X}{S}\} \\
    v_d = \inf_{X \in \{0, 1\}^{\Omega}, i \in \Omega}\{\conv{\indicator{X}}{W}(i) ~|~ i \in \dil{X}{S}\}
\end{align}

Given that $\conv{\indicator{X}}{W} + \conv{\indicator{\bar{X}}}{W} = \sum_{i \in \Omega}{w_i}$, we can write:

\begin{align}
v_e &= \inf_{X \in \{0, 1\}, i \in \Omega}\{\conv{\indicator{X}}{W}(i) ~|~ i \in \ero{X}{S}\}\\
&= \inf_{X \in \{0, 1\}, i \in \Omega}\{\conv{\indicator{\bar{X}}}{W}(i) ~|~ i \in \ero{\bar{X}}{S}\} \\
&= \inf_{X \in \{0, 1\}, i \in \Omega}\{\conv{\indicator{\bar{X}}}{W}(i) ~|~ i \in \overline{\dil{X}{S}}\} \\
&= \inf_{X \in \{0, 1\}, i \in \Omega}\{\conv{\indicator{\bar{X}}}{W}(i) ~|~ i \in \overline{\dil{X}{S}}\} \\
&= \inf_{X \in \{0, 1\}, i \in \Omega}\{\conv{\sum_{i \in \Omega}{w_i} - \indicator{X}}{W}(i) ~|~ i \notin \dil{X}{S}\} \\
&= \sum_{i \in \Omega}{w_i} - \sup_{X \in \{0, 1\}, i \in \Omega}\{\conv{\indicator{X}}{W}(i) ~|~ i \notin \dil{X}{S}\} \\
v_e &= \sum_{i \in \Omega}{w_i} - u_d
\end{align}

The same can be done for the other bound: $u_e = \sum_{i \in \Omega}{w_i - v_d}$.

%% file: appendix.tex
\section{Binary Structuring Element Layer}

One of CNNs' strengths is the ability to learn multiple filters per layer. We also want to be able to learn multiple filters. In CNNs, each final channel is a sum of one filter per input channel. In our case, the final result is either a union or an intersection of the morphological operators (dilation or erosion). Therefore, we want a layer that can learn the union or intersection of any combination of inputs. Let us consider $n$ binary images $x_1, ..., x_n \subset \Omega$. Let $\cC \subset \discint{1}{n}$. Then the intersection and union are given as: 

\begin{align}
\indicator{\bigcap_{i \in \cC}{x_i}} &= \Big( \sum_{i \in \cC}{\indicator{x_i}} \geq \card{\cC}\Big)\\
\indicator{\bigcap_{i \in \cC}{x_i}} &= \Big( \sum_{i \in \cC}{\indicator{x_i}} \geq 1\Big)
\end{align}

As with the BiSE, we can use a single scalar to discriminate between the union or the intersection. To learn the set $\cC$, we can use a parameter $\beta_i$ for each image. This gives the following definition:

\begin{definition}[$\lui$]
 Let $\beta = (\beta_1, ..., \beta_c) \in \R^c$.  Let $\xi$ be a smooth increasing threshold. Let $b \in \R_+$ be a bias and $p \in \R$ a scaling factor. We define the \textit{$\lui$} (Layer Intersection Union) as a thresholded linear combination:
 
 \begin{equation}
   \lui^{\beta}: x \in (\Z^d)^c \mapsto \xi\bigg(p \Big(\sum_{i = 1}^c{\beta_ix_i} - f^+(b) \Big)\bigg) \in \Z^d
 \end{equation}
\end{definition}

A $\lui$ layer can learn any intersection or union of any number of almost binary inputs. We denote $\I = \bigtimes_{k=1}^n{\almostbinary{u_k}{v_k}}$ the set of images with $n$ almost binary channels.

\begin{proposition}[$\lui$ intersection / union equivalence]

Let $n \in \N^*$ and $\cC \subset \discint{1}{n}$. Let $b \in \R$. Let $u_1 < v_1, ..., u_n < v_n \in [0,1]$. Let $\beta \in \R_+^n$. \\

\begin{itemize}
    \item $\lui$ is an intersection by $\cC$ if and only if
    \begin{equation}
        \sum_{k=1}^n{\beta_k} - \min_{k\in \cC}{\Big[(1-u_k)\beta_k\Big]} \leq b < \sum_{k\in \cC}{\beta_k v_k}
    \end{equation}
    
    \item $\lui$ is a union by $\cC$ if and only if 
    \begin{equation}
        \sum_{k \in \cC}{\beta_k u_k} + \sum_{k \in \discint{1}{n} \backslash \cC}{\beta_k} \leq b < \min_{k \in \cC}(\beta_kv_k)
    \end{equation}
    
\end{itemize}

\end{proposition}

Like for the BiSE, if the $\lui$ is properly learned (\textit{i.e. inequalities are respected}), the set $\cC$ can be found by thresholding the $(\beta_i)_{i}$ for a certain value. Moreover, if the inequalities are strict, if all the channels of an input image $I$ are almost binary in $\almostbinary{u}{v}$, then the output $\lui^{\beta,b}(I)$ is almost binary: $\lui^{\beta, b}(I) \in \almostbinary{\xi(u - b)}{\xi(v - b)}$.

We combine the BiSE neurons and the $\lui$ to be able to learn the morphological operators, and aggregate them as unions or intersections.

\begin{definition}[BiSEL]
A BiSEL (BiSE Layer) is the combination of multiple BiSE and multiple $\lui$.
Let $(\bise_{n, k})$ be $N*K$ BiSE and $(\lui_k)_k$ be $K$ $\lui$. Then we define a BiSEL as:

\begin{equation}
\bisel : x \in (\Z^d)^n \mapsto \bigg(\lui_k\Big[\big(\bise_{n, k}(x_n)\big)_n\Big]\bigg)_k \in (\Z^d)^K
\end{equation}

\end{definition}

Given almost binary inputs, the outputs of the BiSEL are also almost binary. 

\begin{figure}[h]
    \centering
    \includegraphics[width=.7\linewidth]{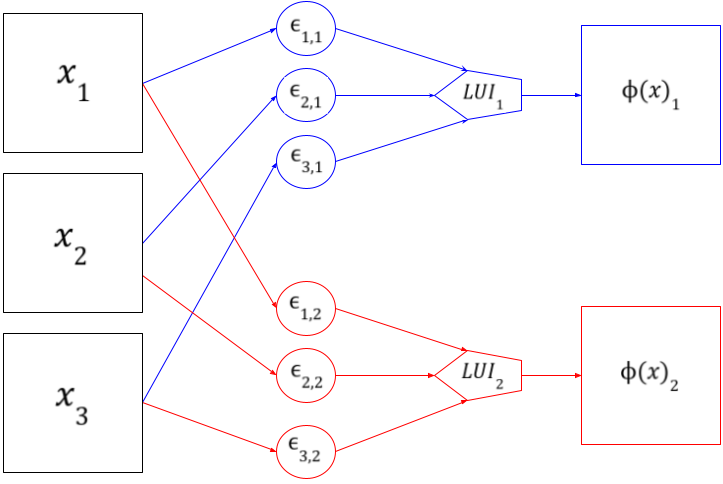}
    \caption{Schema of BiSEL. Input $x$ with 3 input channels. Output $\phi(x)$ with 2 channels.}
    \label{fig:bisel}
\end{figure}

This follows the same logic as CNN. In CNN, at each layer, we have $k$ filters. Each filter applies one convolution to each channel, then we apply a linear combination to all convoluted channels. In BiSEL, we have $K$ filters (the number of $\lui$). For each filter, we apply a morphological operation to a channel, then we aggregate by taking the intersection or union of any channels. See figure \ref{fig:bisel}.

The parameters are the BiSE weights $\{W_{n, k}\}_{n, k} \subset \R^{\Omega}$, the BiSE biases $\{b_{n, k}\}_{n, k} \subset \R$, the $\lui$ parameters $\{\beta_{n, k}\}_{n, k} \subset \R$ and the $\lui$ biases $\{b^l_{n,k}\}_{n, k} \subset \R$. There are $NK(\card{\Omega} + 3)$ parameters.

